\documentclass[11pt,a4paper]{article}
\usepackage[hyperref]{acl2019}
\usepackage{times}
\usepackage{latexsym}
\usepackage{graphicx}

\usepackage{url}
\usepackage{multirow}
\usepackage{dirtytalk}
\usepackage{amsmath}
\usepackage{verbatim}
\usepackage{subcaption}

\usepackage{booktabs}
\newcommand{\ra}[1]{\renewcommand{\arraystretch}{#1}}
\usepackage{arydshln}
\DeclareMathOperator*{\argmax}{argmax} 

\aclfinalcopy 

\usepackage{color}
\usepackage{xspace}

\newcommand{\Loss}[1]{\mathcal{L}_{\mathrm{#1}}}
\newcommand{\Tag}[1]{{{/\texttt{#1}}}}
\newcommand{\TagGreen}[1]{{{/\texttt{\color{green}\textbf{#1}}}}}
\newcommand{\TagRed}[1]{{{/\texttt{\color{red}\textbf{#1}}}}}
\newcommand{\TagBlue}[1]{{{/\texttt{\color{blue}\textbf{#1}}}}}
\newcommand{\ModelName}[1]{{\textsc{#1}}}
\newcommand{\ModelNameText}[1]{{\small\textsc{#1}}}
\newcommand{\ModelNameSmallest}[1]{{\scriptsize\ModelName{#1}}}

\title{Multi-Task Networks With Universe, Group, and Task Feature Learning}

\author{Shiva Pentyala\thanks{\ \ \ This work was done while Shiva Pentyala was interning at Amazon Alexa.} \\
  Texas A\&M University \\
  {\tt pk123@tamu.edu } \\\And
  Mengwen Liu \\
  Amazon Alexa \\
  {\tt mengwliu@amazon.com} \\\And
  Markus Dreyer \\
  Amazon Alexa \\
  {\tt mddreyer@amazon.com} \\}

\date{}

\begin{document}
\maketitle

\begin{abstract}


  We present methods for multi-task learning that take advantage of natural
  groupings of related tasks. Task groups may be defined along known
  properties of the tasks, such as task domain or language. Such task groups
  represent supervised information at the inter-task level and can be encoded
  into the model. We investigate two variants of neural network architectures
  that accomplish this, learning different feature spaces at the levels of
  individual tasks, task groups, as well as the universe of all tasks: (1)    parallel architectures encode each input \emph{simultaneously} into feature spaces at different levels; (2) serial architectures encode each input \emph{successively} into feature spaces at different levels in the task hierarchy. We demonstrate the methods on natural language understanding (NLU)
  tasks, where a grouping of tasks into different task domains leads to
  improved performance on ATIS, Snips, and a large in-house dataset.
  
\end{abstract}

\section{Introduction}

In multi-task learning \cite{Caruana1993}, multiple related tasks are learned
together. Rather than learning one task at a time, multi-task learning uses
information sharing between multiple tasks. This technique has been shown to
be effective in multiple different areas, e.g., vision
\cite{Zhang2014FacialLD}, medicine \cite{Bickel2008MultitaskLF}, and natural
language processing \cite{collobert2008unified,luong2016,Fan2017TransferLF}.

The selection of tasks to be trained together in multi-task learning can be
seen as a form of supervision: The modeler picks tasks that are known \emph{a
  priori} to share some commonalities and decides to train them together. In
this paper, we consider the case when information about the
\emph{relationships} of these tasks is available as well, in the form of
natural \emph{groups} of these tasks. Such task groups can be available in
various multi-task learning scenarios: In \emph{multi-language} modeling, when
learning to parse or translate multiple languages jointly, information on
language families would be available; in \emph{multimodal} modeling, e.g.,
when learning text tasks and image tasks jointly, clustering the tasks into
these two groups would be natural. In \emph{multi-domain} modeling, which is
the focus of this paper, different tasks naturally group into different
domains.

We hypothesize that adding such inter-task supervision can encourage a
model to generalize along the desired task dimensions. We introduce
neural network architectures that can encode task groups, in two
variants:

\begin{itemize}
\item Parallel network architectures encode each input
  \emph{simultaneously} into feature spaces at different levels;
\item serial network architectures encode each input
  \emph{successively} into feature spaces at different levels in the
  task hierarchy.
\end{itemize}

These neural network architectures are general and can be applied to
any multi-task learning problem in which the tasks can be grouped into
different task groups.

\paragraph{Application Example.}
To illustrate our method, we now introduce the specific scenario that we use
to evaluate our method empirically: multi-domain natural language
understanding (NLU) for virtual assistants. Such assistants, e.g., Alexa,
Cortana, or Google Assistant, perform a range of tasks in different domains
(or, categories), such as \emph{Music}, \emph{Traffic}, \emph{Calendar},
etc. With the advent of frameworks like Alexa Skills Kit, Cortana Skills Kit
and Actions on Google, third-party developers can extend the capabilities of
these virtual assistants by developing new tasks, which we call \emph{skills},
e.g., \emph{Uber}, \emph{Lyft}, \emph{Fitbit}, in any given domain. Each skill
is defined by a set of intents that represents different functions to handle a
user's request, e.g., \texttt{play\_artist} or \texttt{play\_station} for a
skill in the \textit{Music} domain. Each intent can be instantiated with
particular slots, e.g., \texttt{artist} or \texttt{song}. An utterance like ``play
madonna'' may be parsed into intent and slots, resulting in a structure like
\texttt{PlayArtist(artist="madonna")}. Skill developers provide their own
labeled sample utterances in a grammar-like format \cite{ask2017},
individually choosing a label space that is suitable for their
problem.\footnote{They may choose to pick from a set of predefined labels with
  prepopulated content, e.g., \emph{cities} or \emph{first names}.} We learn
intent classification (IC) and slot filling (SF) models for these skills, in
order to recognize user utterances spoken to these skills that are similar but
not necessarily identical to the sample utterances given by their skill
developers.

In this paper, we apply multi-task learning to this problem, learning the
models for all skills jointly, as the individual training data for any given
skill may be small and utterances across multiple skills have similar
characteristics, e.g., they are often short commands in spoken language. In
addition, we wish to add information on task groups (here: skill domains) into
these models, as utterances in different skills of the same domain may be
especially similar. For example, although the \emph{Uber} and \emph{Lyft}
skills are built by independent developers, end users may speak similar
utterances to them. Users may say ``get me a ride'' to
\emph{Uber} and ``I'm requesting a ride'' to \emph{Lyft}.

\paragraph{Contributions.} The main contributions of this paper are as follows:
\begin{itemize}
\item We introduce unified models for multi-task learning that learn three
  sets of features: \emph{task}, \emph{task group}, and \emph{task universe} features;
\item we introduce two architecture variants of such multi-task models:
  \emph{parallel} and \emph{serial} architectures;
\item we evaluate the proposed models to perform multi-domain joint learning
  of slot filling and intent classification on both public datasets and
  a real-world dataset from the Alexa virtual assistant;
\item we demonstrate experimentally the superiority of introducing group-level
  features and learning features in both parallel and serial ways.
\end{itemize}

\section{Proposed Architectures}
The goal of multi-task learning (MTL) is to utilize shared information across related tasks. The features learned in one task could be transferred to reinforce the feature learning of other tasks, thereby boosting the performance of all tasks via mutual feedback within a unified MTL architecture. We consider the problem of multi-domain natural language understanding (NLU) for virtual assistants. Recent progress has been made to build NLU models to identify and extract structured information from user's request by jointly learning intent classification (IC) and slot filling (SF) \cite{tur2010left}. However, in practice, a common issue when building NLU models for every skill is that the amount of annotated training data varies across skills and is small for many individual skills. Motivated by the idea of learning multiple tasks jointly, the paucity of data can be resolved by transferring knowledge between different tasks that can reinforce one another.


In what follows, we describe four end-to-end MTL architectures
(Sections~\ref{method:parallel} to \ref{method:serial-3}). These
architectures are encoder-decoder architectures where the encoder
extracts three different sets of features: \emph{task},
\emph{task group}, and \emph{task universe} features, and the decoder
produces desired outputs based on feature representations. In
particular, the first one (Figure~\ref{figure:parallel}) is a parallel
MTL architecture where task, task group, and task universe features
are encoded in parallel and then concatenated to produce a composite
representation. The next three architectures
(Figure~\ref{figure:serial}) are serial architectures in different
variants: In the first serial MTL architecture, group and universe
features are learned first and are then used as inputs to learn
task-specific features. The next serial architecture is similar but
introduces highway connections that feed representations from earlier
stages in the series directly into later stages. In the last architecture,
the order of serially learned features is changed, so that task-specific
features are encoded first.

In Section \ref{method:single}, we introduce an encoder-decoder
architecture to perform slot filling and intent classification jointly
in a multi-domain scenario for virtual assistants. Although we conduct experiments on multi-domain NLU systems of
virtual assistants, the architectures can easily be applied to other tasks. Specifically, the encoder/decoder could be instantiated with any components or architectures, i.e., Bi-LSTM \cite{hochreiter1997long} for the encoder, and classification or sequential labeling for the decoder. 


\subsection{Parallel MTL Architecture}
\label{method:parallel}

\begin{figure}
  \centering
    \includegraphics[scale=0.7,width=\columnwidth]{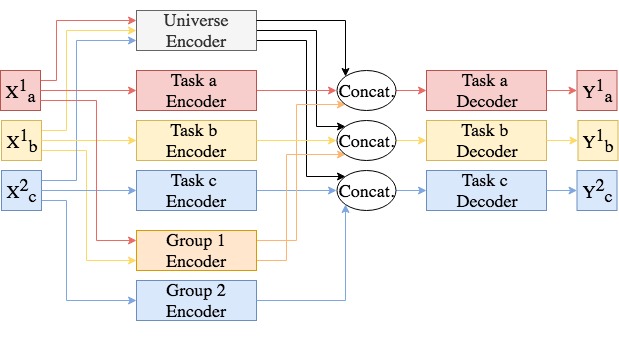} %
	\caption{The \ModelNameText{Parallel[Univ+Group+Task]} architecture, which learns universe, group, and task features. Three tasks $a$, $b$, and $c$ are illustrated in the figure where \mbox{$a, b \in \text{group}_1$} and $c \in \text{group}_2$.}
	\label{figure:parallel}
\end{figure}

The first architecture, shown in Figure \ref{figure:parallel}, is designed to learn the three sets of features at the same stage; therefore we call it a parallel MTL architecture, or \ModelNameText{Parallel[Univ+Group+Task]}. This architecture uses three types of encoders: 1) A universe encoder to extract the common features across all tasks; 2) task-specific encoders to extract task-specific features; and 3) group-specific encoders to extract features within the same group. Finally, these three feature representations are concatenated and passed through the task-specific decoders to produce the output. 

Assume we are given a MTL problem with $m$ groups of tasks. Each task is associated with a dataset of training examples $D=\{(\boldsymbol{x}_1^i, \boldsymbol{y}_1^i),...,(\boldsymbol{x}_{m_i}^{i}, \boldsymbol{y}_{m_i}^{i})\}_{i=1}^m$, where $\boldsymbol{x}_{k}^i$ and $\boldsymbol{y}_{k}^i$ denote input data and corresponding labels for task $k$ in group $i$. The parameters of the parallel MTL model (and also for the other MTL models) are trained to minimize the weighted sum of individual task-specific losses that can be computed as:
\begin{equation}
\Loss{tasks} = \sum_{i=1}^m \sum_{j=1}^{m_i} \alpha_j^i * \mathcal{L}_j^i,
\label{eq:loss_task}
\end{equation}
where $\alpha_j^i$ is a static weight for task $j$ in group $i$, which could be proportional to the size of training data of the task. The loss function $\mathcal{L}_j^i$ is defined based on the tasks performed by the decoder, which will be described in Section~\ref{method:single}.

To eliminate redundancy in the features learned by three different types of encoders, we add adversarial loss and orthogonality constraints cost ~\cite{bousmalis2016domain,liu2017adversarial}. Adding adversarial loss aims to prevent task-specific features from creeping into the shared space. We apply adversarial training to our shared encoders, i.e., the universe and group encoders. To encourage task, group, and universe encoders to learn features from different aspects of the inputs, we add orthogonality constraints between task and universe/group representations of each domain. The loss function defined in Equation \ref{eq:loss_task} becomes:
\begin{equation}
\Loss{all} = \Loss{tasks} + \lambda * \Loss{adv} + \gamma * \Loss{ortho}
\label{eq:loss_all}
\end{equation}
where $\Loss{adv}$ and $\Loss{ortho}$ denote the loss function for adversarial training and orthogonality constraints respectively, and $\lambda$ and $\gamma$ are hyperparameters.

\subsection{Serial MTL Architecture}
\label{method:serial-1}
\begin{figure}[t]
    \centering
    \begin{subfigure}{\columnwidth}
        \includegraphics[width=\columnwidth]{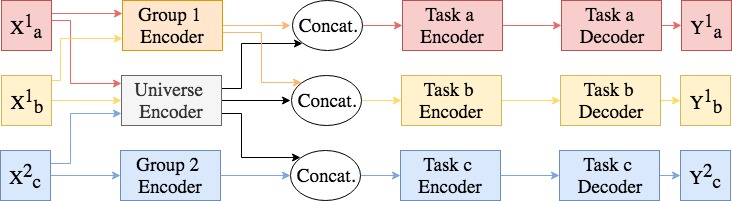}
        \caption{\ModelNameText{Serial}}
        \label{figure:serial-1}
    \end{subfigure}
    \hfill 
    \begin{subfigure}{\columnwidth}
        \includegraphics[width=\columnwidth]{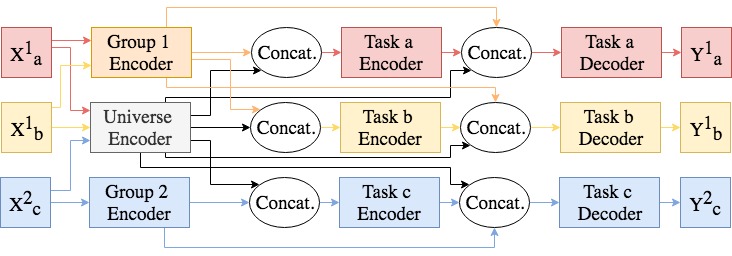}
        \caption{\ModelNameText{Serial+Highway}}
        \label{figure:serial-2}
    \end{subfigure}
	\hfill
    \begin{subfigure}{\columnwidth}
        \includegraphics[width=\columnwidth]{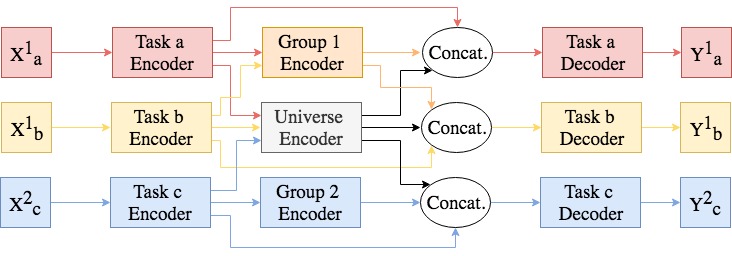}
        \caption{\ModelNameText{Serial+Highway+Swap}}
        \label{figure:serial-3}
    \end{subfigure}
    \caption{Three serial MTL architectures. In each of these architectures, individual decoders utilize all three sets of features (task, universe, and group features) to perform a task. Three tasks $a$, $b$, and $c$ are illustrated in the figures where \mbox{$a, b \in \text{group}_1$} and $c \in \text{group}_2$.}\label{figure:serial}
\end{figure}

The second MTL architecture, called \ModelNameText{Serial}, has the same set of encoders and decoders
as the parallel MTL architecture. The differences are 1) the order of
learning features and 2) the input for individual decoders.  In this
serial MTL architecture, three sets of features are learned in a
sequential way in two stages. As shown in Figure \ref{figure:serial-1}, group
encoders and a universe encoder encode group-level and
fully shared universe-level features, respectively, based on input data. Then, task encoders use that concatenated feature representation to learn task-specific features. Finally, in this serial architecture, the individual task decoders use their corresponding private encoder outputs only to perform tasks. This contrasts with the parallel MTL architecture, which uses combinations of three feature representations as input to their respective task decoders. 

\subsection{Serial MTL Architecture with Highway Connections}
\label{method:serial-2}
Decoders in the \ModelNameText{Serial} architecture, introduced in the previous section,
do not have direct access to group and universe feature representations. However, directly utilizing these shared features could be beneficial for some tasks. Therefore, we add highway connections to incorporate universe encoder output and corresponding group encoder outputs as inputs to the individual decoders in addition to task-specific encoder output; we call this model \ModelNameText{Serial+Highway}. 

As shown in Figure \ref{figure:serial-2}, input to the task-specific encoders are the same as those in the serial MTL architecture, i.e., the concatenation of the group and universe features. The input to each task-specific decoder, however, is now the concatenation of the features from the group encoder, the universe encoder, and the task-specific encoder. 



\subsection{Serial MTL Architecture with Highway Connections and Feature Swapping}
\label{method:serial-3}

In both serial MTL architectures introduced in the previous two sections, the input to the task encoders is the output of the more general group and universe encoders. That output potentially underrepresents some task-specific aspects of the input. Therefore, we introduce \ModelNameText{Serial+Highway+Swap}; a variant of \ModelNameText{Serial+Highway}, in which the two stages of universe/group features and task-specific features are swapped. As shown in Figure \ref{figure:serial-3}, the task-specific
representations are now learned in the first stage, and group and universe
feature representations based on the task features are learned in the
second stage. In this model, the task encoder directly takes input data and learns task-specific features. Then, the universe encoder and group encoders take the task-specific representations as input and generate fully shared universe and group-level representations, respectively. Finally, task-specific decoders use the concatenation of all three features -- universe, group and task features, to perform the final tasks. 



\subsection{An Example of Encoder-Decoder Architecture for a Single Task}
\label{method:single}

All four MTL architectures introduced in the previous sections are general such that they could be applied to many applications. In this section, we use the task of joint slot filling (SF) and intent classification (IC) for natural language understanding (NLU) systems for virtual assistants as an example. We design an encoder-decoder architecture to perform SF and IC as a joint task, on top of which the four MTL architectures are built. 

\begin{figure}[t]
\begin{center}
\includegraphics[width=\columnwidth]{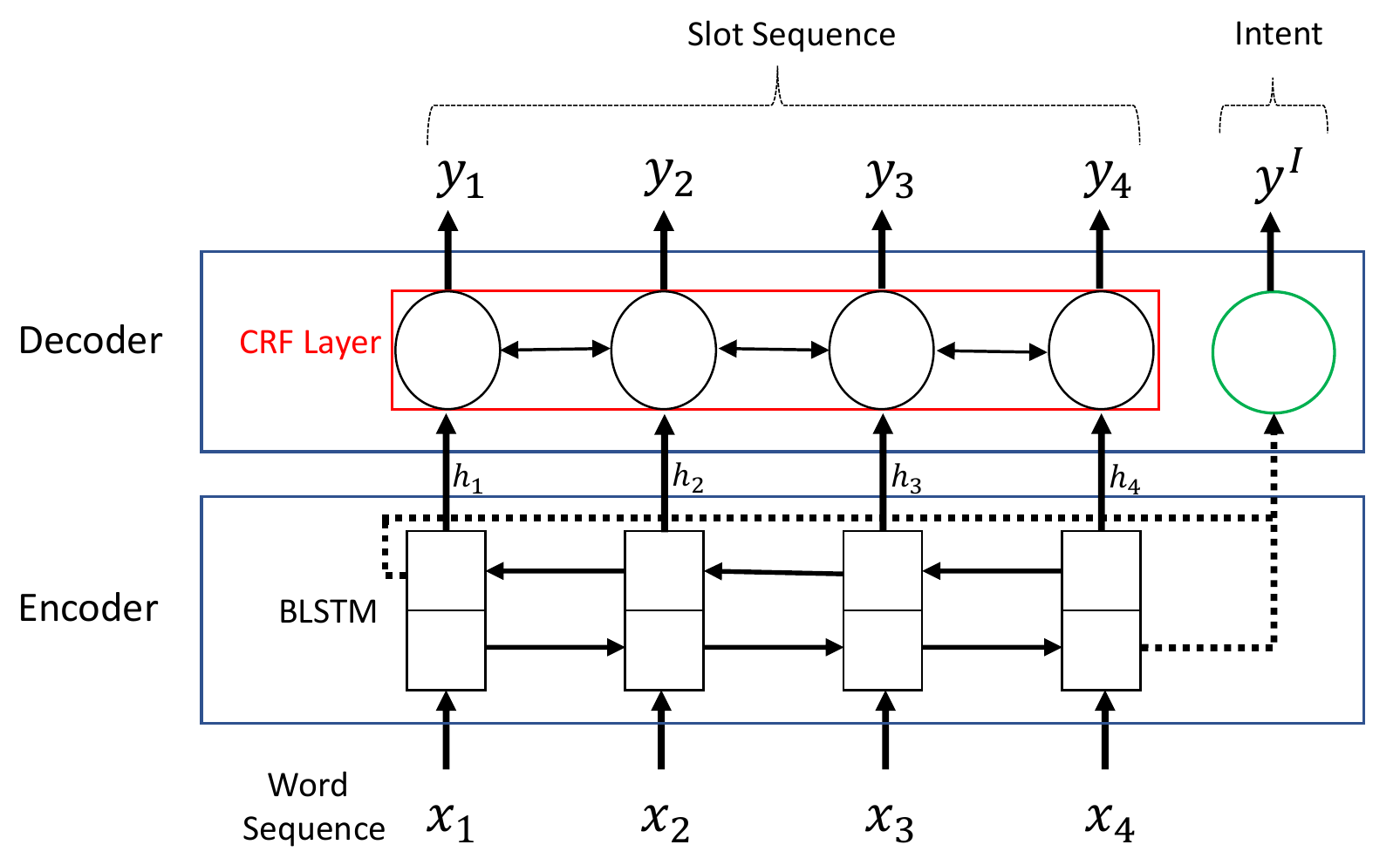}
\caption{\ModelName{Joint-SF-IC} model.}
\label{fig:joint-sf-ic}
\end{center}
\end{figure}

Given an input sequence $\boldsymbol{x}=(x_1,\ldots,x_T)$, the goal is to jointly learn an equal-length tag sequence of slots $\boldsymbol{y}^S = (y_1,\ldots,y_T)$ and the overall intent label $y^I$. By using a joint model, rather than two separate models, for SF and IC, we exploit the correlation of the two output spaces. For example, if the intent of a sentence is \texttt{book\_ride} it is likely to contain the slot types \texttt{from\_address} and \texttt{destination\_address}, and vice versa. The \ModelNameText{Joint-SF-IC} model architecture is shown in Figure~\ref{fig:joint-sf-ic}. It is a simplified version compared to the \ModelNameText{SlotGated} model \cite{goo2018slot}, which showed state-of-the-art results in jointly modeling SF and IC. Our architecture uses neither slot/intent attention nor a slot gate.

To address the issues of  small amounts of training data and out-of-vocabulary (OOV) words, we use character embeddings, learned during training, as well as pre-trained word embeddings \cite{lample2016neural}. These word and character representations are passed as input to the encoder, which is a bidirectional long short-term memory (Bi-LSTM) \cite{hochreiter1997long} layer that computes forward hidden state $\overrightarrow{\boldsymbol{h}_t}$ and backward hidden state $\overleftarrow{\boldsymbol{h_t}}$ per time step $t$ in the input sequence. We then concatenate $\overrightarrow{\boldsymbol{h}_t}$  and $\overleftarrow{\boldsymbol{h}_t}$ to get final hidden state $\boldsymbol{h}_t = [\overrightarrow{\boldsymbol{h}_t};\overleftarrow{\boldsymbol{h}_t}]$ at time step $t$. 

\noindent \textbf{Slot Filling (SF)}: For a given sentence $\boldsymbol{x}=(x_1,\ldots,x_T)$ with $T$ words, we use their respective hidden states $\boldsymbol{h}=(\boldsymbol{h}_1,\ldots,\boldsymbol{h}_T)$ from the encoder (Bi-LSTM layer) to model tagging decisions $\boldsymbol{y}^S=(y_1,\ldots,y_T)$ jointly using a conditional random field (CRF) layer \cite{lample2016neural,lafferty2001conditional}: 
\begin{equation}
\boldsymbol{y}^S = \argmax_{\boldsymbol{y} \in \mathcal{Y^S}}  f_S(\boldsymbol{h}, \boldsymbol{x}, \boldsymbol{y}),
\label{eq:slot}       
\end{equation}
where $\mathcal{Y^S}$ is the set of all possible slot sequences, and $f_S$ is the CRF decoding function.

\noindent \textbf{Intent Classification (IC)}: Based on the hidden states from the encoder (Bi-LSTM layer), we use the last forward hidden state $\overrightarrow{\boldsymbol{h}_T}$ and last backward hidden state $\overleftarrow{\boldsymbol{h}_1}$ to compute the moment $\boldsymbol{h}^I = [\overrightarrow{\boldsymbol{h}_T};\overleftarrow{\boldsymbol{h}_1}]$ which can be regarded as the representation of the entire input sentence. Lastly, the intent $y^I$ of the input sentence is predicted by feeding $\boldsymbol{h}^I$ into a fully-connected layer with softmax activation function to generate the prediction for each intent:
\begin{equation}
y^I = \text{softmax}(\boldsymbol{W}^I_{hy} \cdot \boldsymbol{h}^I+b),
\label{eq:intent}       
\end{equation}
where $y^I$ is the prediction label, $\boldsymbol{W}^I_{hy}$ is a weight matrix and $b$ is a bias term. 


\noindent \textbf{Joint Optimization}: As our decoder models a joint task of SF and IC, we define the loss $\mathcal{L}$ as a weighted sum of individual losses which can be plugged into $\mathcal{L}_j^{i}$ in Equation \ref{eq:loss_task}:
\begin{equation}
\Loss{task}=w_\text{SF}*\Loss{SF} + w_\text{IC}*\Loss{IC},
\label{eq:task-sf-ic}
\end{equation}
where $\Loss{SF}$ is the cross-entropy loss based on the probability of the correct tag sequence \cite{lample2016neural}, $\Loss{IC}$ is the cross-entropy loss based on the predicted and true intent distributions \cite{liu2017adversarial} and $w_\text{SF}$, $w_\text{IC}$ are hyperparameters to adjust the weights of the two loss components.

\section{Experimental Setup}

\subsection{Dataset}

We evaluate our proposed models for multi-domain joint slot filling and intent classification for spoken language understanding systems. We use the following benchmark dataset and large-scale Alexa dataset for evaluation, and we use classic intent accuracy and slot F1 as in \newcite{goo2018slot} as evaluation metrics.

\begin{table}[!htbp]
\centering
\ra{1.2}
\fontsize{9}{9}\selectfont
\begin{tabular}{lrr}\toprule
\textbf{Property} & \textbf{ATIS} & \textbf{Snips} \\ \hline
Train set size    & 4,478          & 13,084          \\ 
Dev set size      & 500           & 700            \\ 
Test set size     & 893           & 700            \\ \hdashline
\#Slots           & 120           & 72             \\ 
\#Intents         & 21            & 7              \\ \bottomrule
\end{tabular}
\caption{Statistics of the benchmark dataset.}
\label{table:benchmark data}
\end{table}

\noindent \textbf{Benchmark Dataset}: We consider two widely used datasets ATIS~\cite{tur2010left} and Snips \cite{goo2018slot}. The statistics of these datasets are shown in Table \ref{table:benchmark data}. For each dataset, we use the same train/dev/test set as \newcite{goo2018slot}. ATIS is a single-domain (Airline Travel) dataset while Snips is a more complex multi-domain dataset due to the intent diversity and large vocabulary. 

For initial experiments, we use ATIS and Snips as two tasks. For multi-domain experiments, we split Snips into three domains -- \emph{Music}, \emph{Location}, and \emph{Creative} based on its intents and treat each one as an individual task. Thus for this second set of experiments, we have four tasks (ATIS and Snips splits). Table \ref{table:snips split} shows the new datasets obtained by splitting Snips. This new dataset allows us to introduce task groups. We define ATIS and Snips-location as one task group, and Snips-music and Snips-creative as another. 

\begin{table}[b]
\centering
\ra{1.2}
\fontsize{9}{9}\selectfont
\begin{tabular}{ll}
\toprule
\textbf{Dataset} & \textbf{Intent}                                                                                   \\ \hline
Snips-creative   & \begin{tabular}[l]{@{}l@{}}\texttt{search\_creative\_work}\\\texttt{rate\_book}\end{tabular}                        \\ \hdashline
Snips-music      & \begin{tabular}[l]{@{}l@{}}\texttt{play\_music}\\\texttt{add\_to\_playlist}\end{tabular}                            \\ \hdashline
Snips-location   & \begin{tabular}[l]{@{}l@{}}\texttt{get\_weather}\\\texttt{book\_restaurant}\\\texttt{search\_screening\_event}\end{tabular} \\ \bottomrule
\end{tabular}
\caption{Snips after splitting based on intent.}
\label{table:snips split}
\end{table}

\begin{table}[!htbp]
\centering
\ra{1.2}
\fontsize{9}{9}\selectfont
\begin{tabular}{lrr}
\toprule
\multirow{2}{*}{\textbf{Domain\slash Group}} & \multicolumn{2}{c}{\textbf{Skill Count}} \\ \cmidrule(lr){2-3}
                                 & Train   & Dev   \\ \hline
Games, Trivia \& Accessories     & 37                   & 37                 \\ 
Smart Home                       & 12                   & 4                  \\ 
Music \& Audio                   & 8                    & 8                  \\ 
Lifestyle                        & 7                    & 7                  \\ 
Education \& Reference           & 7                    & 7                  \\ 
Novelty \& Humor                 & 6                    & 6                  \\ 
Health \& Fitness                & 5                    & 5                  \\ 
Food \& Drink                    & 3                    & 3                  \\ 
Movies \& TV                     & 3                    & 3                  \\ 
News                             & 2                    & 0                  \\ \hdashline
\textbf{Total}                   & \textbf{90}          & \textbf{80}        \\ \bottomrule
\end{tabular}
\caption{Statistics of the Alexa dataset.}
\label{table:alexa data}
\end{table}

\noindent \textbf{Alexa Dataset}: We use live utterances spoken to 90 Alexa skills with the highest traffic. These are categorized into 10 domains, based on assignments by the developers of the individual skills. Each skill is a task in the MTL setting, and each domain acts as a task group. Due to the limited annotated datasets for skills, we do not have validation sets for these 90 skills. Instead, we use another 80 popular skills that fall into the same domain groups as the 90 skills as the validation set to tune model parameters. Table \ref{table:alexa data} shows the statistics of the Alexa dataset based on domains. For training and validation sets, we keep approximately the same number of skills per group to make sure that hyperparameters of adversarial training are unbiased. We use the validation datasets to choose the hyperparameters for the baselines as well as our proposed models. 





\subsection{Baselines}

We compare our proposed model with the following three competitive architectures for single-task joint slot filling (SF) and intent classification (IC), which have been widely used in prior literature: 
\begin{itemize}
\item \ModelNameText{JointSequence:} \newcite{hakkani2016multi} proposed a Bi-LSTM joint model for slot filling, intent classification, and domain classification.
\item \ModelNameText{AttentionBased:} \newcite{liu2016attention} showed that incorporating an attention mechanism into a Bi-LSTM joint model can reduce errors on intent detection and slot filling.
\item \ModelNameText{SlotGated:} \newcite{goo2018slot} added a slot-gated mechanism into the traditional attention-based joint architecture, aiming to explicitly model the relationship between intent and slots, rather than implicitly modeling it with a joint loss.
\end{itemize}

We also compare our proposed model with two closely related multi-task learning (MTL) architectures that can be treated as simplified versions of our parallel MTL architecture:
\begin{itemize}
\item \ModelNameText{Parallel[Univ]:} This model, proposed by \newcite{liu2017adversarial}, uses a universe encoder that is shared across all tasks, and decoders are task-specific.
\item \ModelNameText{Parallel[Univ+Task]:} This model, also proposed by \newcite{liu2017adversarial}, uses task-specific encoders in addition to the shared encoder. To ensure non-redundancy in features learned across shared and task-specific encoders, adversarial training and orthogonality constraints are incorporated.
\end{itemize}


\subsection{Training Setup}

All our proposed models are trained with backpropagation, and gradient-based optimization is performed using Adam \cite{kingma2014adam}. In all experiments, we set the character LSTM hidden size to 64 and word embedding LSTM hidden size to 128. We use 300-dimension GloVe vectors \cite{pennington2014glove} for the benchmark datasets and in-house embeddings for the Alexa dataset, which are trained with Wikipedia data and live utterances spoken to Alexa. Character embedding dimensions and dropout rate are set to 100 and 0.5 respectively. Minimax optimization in adversarial training was implemented via the use of a gradient reversal layer \cite{ganin2015unsupervised,liu2017adversarial}. The models are implemented with the TensorFlow library \cite{abadi2016tensorflow}. 

For benchmark data, the models are trained using an early-stop strategy with
maximum epoch set to 50 and patience (i.e., number of epochs with no
improvement on the dev set for both SF and IC) to 6. In addition, the benchmark dataset has varied size vocabularies across its datasets. To give equal importance to each of them, $\alpha_i^j$ (see Equation~\ref{eq:loss_task}) is proportional to $1/n$, where $n$ is the training set size of task $j$ in group $i$. We are able to train on CPUs, due to the low values of $n$.

For Alexa data, optimal hyperparameters are determined on the 80 development skills and applied to the training and evaluation of the 90 test skills. $\alpha_i^j$ is here set to 1 as all skills have $10,000$ training utterances sampled from the respective developer-defined skill grammars \cite{ask2017}. Here, training was done using GPU-enabled EC2 instances (p2.8xlarge). 

Our detailed training algorithm is similar to the one used by \newcite{collobert2008unified} and \newcite{liu2016recurrent,liu2017adversarial}, where training is achieved in a stochastic manner by looping over the tasks. For example, an epoch involves these four steps: 1) select a random skill; 2) select a random batch from the list of available batches for this skill; 3) update the model parameters by taking a gradient step w.r.t this batch; 4) update the list of available batches for this skill by removing the current batch.

\section{Experimental Results}

\begin{table}[!t]
	\centering
        \ra{1.2}
	\fontsize{7}{7}\selectfont
	\begin{tabular}{lcccc}\toprule
				\multirow{3}{*}{\textbf{Model}} & \multicolumn{2}{c}{\textbf{ATIS}} & \multicolumn{2}{c}{\textbf{Snips}}  \\
                                \cmidrule(lr){2-3}
                                \cmidrule(lr){4-5}
				\multicolumn{1}{c}{} & Intent & Slot & Intent & Slot \\
				\multicolumn{1}{c}{} & Acc. & F1 & Acc. & F1 \\
				\cline{1-5}
				\ModelName{JointSequence} & 92.6 & 94.3 & 96.9 & 87.3 \\
				\ModelName{AttentionBased} & 91.1 & 94.2 & 96.7 & 87.9 \\
	            \ModelName{SlotGated} & 93.6 & 94.8 & 97.0 & 88.8 \\
	            \ModelName{Joint-SF-IC} & 96.1 & 95.4 & \textbf{98.0} & \textbf{94.8} \\
				\cdashline{1-5}
	            \ModelName{Parallel[Univ]} & 95.9 & 95.1 & 98.1 & 94.3 \\
	            \ModelName{Parallel[Univ+Task]} & \textbf{96.6} & \textbf{95.8} & 97.6 & 94.5 \\
	            \bottomrule
	        \end{tabular}
	\caption{Results on benchmark datasets (ATIS and original Snips).} 
	\label{table:result_benchmark_1}
\end{table}

\begin{table}[!t] 
	\fontsize{7}{7}\selectfont
	\centering
        \ra{1.2}
	\begin{tabular}{lcccc}\toprule
				\multirow{3}{*}{\textbf{Model}} & \multicolumn{2}{c}{\textbf{ATIS}} & \multicolumn{2}{c}{\textbf{Snips-location}}  \\
                                \cmidrule(lr){2-3}
                                \cmidrule(lr){4-5}
				\multicolumn{1}{c}{} & Intent & Slot & Intent & Slot \\
				\multicolumn{1}{c}{} & Acc. & F1 & Acc. & F1 \\
				\cline{1-5}
	                        \ModelName{Joint-SF-IC} & 96.1 & 95.4 & 99.7 & 96.3 \\
				\cdashline{1-5}
				\ModelName{Parallel[Univ]} & 96.4 & 95.4 & 99.7 & 95.8  \\
				\ModelName{Parallel[Univ+Task]} & 96.2 & 95.5 & 99.7 & 96.0 \\
	                        \ModelName{Parallel[Univ+Group+Task]} & 96.9 & 95.4 & 99.7 & 96.5  \\
				\hdashline
	                        \ModelName{Serial} & 97.2 & \textbf{95.8} & \textbf{100.0} &  96.5 \\
				\ModelName{Serial+Highway} & 96.9 & 95.7 & \textbf{100.0} & \textbf{97.2} \\
	                        \ModelName{Serial+Highway+Swap} & \textbf{97.5} & 95.6 & 99.7 & 96.0 \\
	\end{tabular}
	\newline
	\vspace*{0.3 cm}
	\newline
	\begin{tabular}{lcccc}
				\multirow{3}{*}{\textbf{Model}} & \multicolumn{2}{c}{\textbf{Snips-music}} & \multicolumn{2}{c}{\textbf{Snips-creative}}  \\
                                \cmidrule(lr){2-3}
                                \cmidrule(lr){4-5}
				\multicolumn{1}{c}{} & Intent & Slot & Intent & Slot \\
				\multicolumn{1}{c}{} & Acc. & F1 & Acc. & F1 \\
				\cline{1-5}
	                        \ModelName{Joint-SF-IC} & \textbf{100.0} & 93.1 & 100.0 & 96.6\\
				\cdashline{1-5}
				\ModelName{Parallel[Univ]} & \textbf{100.0} & 92.1 & 100.0 & 95.8 \\
				\ModelName{Parallel[Univ+Task]} & \textbf{100.0} & 93.4 & 100.0 & 97.2 \\
	                        \ModelName{Parallel[Univ+Group+Task]} & 99.5 & 94.4 & 100.0 & 97.3\\
				\hdashline
	                        \ModelName{Serial} & \textbf{100.0} & 93.8 & 100.0 & 97.2\\
				\ModelName{Serial+Highway} & 99.5 & \textbf{94.8} & 100.0 & 97.2\\
	                        \ModelName{Serial+Highway+Swap} & \textbf{100.0} & 93.9 & 100.0 & \textbf{97.8}\\
				\bottomrule
	\end{tabular}
	\caption{Results on benchmark dataset (ATIS and subsets of Snips).} 
	\label{table:result_benchmark_2}
\end{table}

\begin{table}[t]
	\centering
        \ra{1.2}
        \resizebox{\columnwidth}{!}{%
	\begin{tabular}{lcccc}\toprule
				\multirow{2}{*}{\textbf{Model}} & \multicolumn{2}{c}{Intent Acc.}  & \multicolumn{2}{c}{Slot F1}  \\
                                \cmidrule(lr){2-3}
                                \cmidrule(lr){4-5}
				\multicolumn{1}{c}{} & Mean & Median & Mean & Median\\
				\cline{1-5}
	                        \ModelName{Joint-SF-IC} & 93.36 & 95.90 & 79.97 & 85.23 \\
				\cdashline{1-5}
				\ModelName{Parallel[Univ]} & \textbf{93.44} & 95.50 & \textbf{80.76} & 86.18 \\
				\ModelName{Parallel[Univ+Task]} & 93.78 & 96.35 & \textbf{80.49} & 85.81 \\
				\ModelName{Parallel[Univ+Group+Task]} & 93.87 & 96.31 & \textbf{80.84} & 86.21 \\
				\cdashline{1-5}
	                        \ModelName{Serial} & 93.83 & 96.24 & \textbf{80.84} & 86.14 \\
				\ModelName{Serial+Highway} & \textbf{93.81} & 96.28 & \textbf{80.73} & 85.71 \\
	                        \ModelName{Serial+Highway+Swap} & \textbf{94.02} & \textbf{96.42} & \textbf{80.80} & \textbf{86.44} \\
	            \bottomrule
	        \end{tabular}
        }
	\caption{Results on the Alexa dataset. Best results on mean intent accuracy and slot F1 values, and results that are not statistically different from the best model are marked in bold.}
	\label{table:result_alexa}
\end{table}

\begin{table*}[t]
        \resizebox{\textwidth}{!}{%
	\centering
        \ra{1.2}
	\begin{tabular}{lcccccccccc}
				\toprule
				\textbf{Model} & \textbf{Education} & \textbf{Food} & \textbf{Games} & \textbf{Health} & \textbf{Lifestyle} & \textbf{Movie} & \textbf{Music} & \textbf{News} & \textbf{Novelty} & \textbf{Smart Home} \\ 
				\hline
	                        \ModelName{Joint-SF-IC} & 95.89 & 90.60 & 96.29 & 92.80 & 93.84 & 67.50 & 93.51 & 90.05 & 95.00 & 89.58 \\
				\hdashline
				\ModelName{Parallel[Univ]} & 95.56 & 89.47 & 95.96 & 90.74 & 94.49 & \textbf{74.40} & 93.29 & \textbf{90.90} & 94.58 & 90.63 \\
				\ModelName{Parallel[Univ+Task]} & 95.99 & 91.10 & 96.50 & 93.60 & 94.46 & 68.40 & 94.45 & 88.60 & 94.93 & 90.66 \\
				\ModelName{Parallel[Univ+Group+Task]} & 96.17 & 91.23 & \textbf{96.58} & 93.92 & 94.33 & 68.10 & 94.56 & 87.15 & 95.00 & 91.06 \\
				\hdashline
	                        \ModelName{Serial} & 96.11 & 90.77 & 96.44 & 94.04 & \textbf{94.63} & 69.07 & 94.59 & 87.30 & 94.92 & 90.89 \\
				\ModelName{Serial+Highway} & 96.04 & 91.70 & 96.45 & 94.10 & 92.71 & 68.67 & 94.86 & 87.90 & 95.03 & 91.41 \\
	                        \ModelName{Serial+Highway+Swap} & \textbf{96.20} & \textbf{91.80} & 96.49 & \textbf{94.16} & 94.37 & 68.37 & \textbf{94.94} & 88.35 & \textbf{95.08} & \textbf{91.64} \\
	            \bottomrule
	        \end{tabular}
        }
	\caption{Intent accuracy on different groups of the Alexa dataset.}
	\label{table:result_alexa_intent_group}
\end{table*}

\begin{table*}[!t]
        \resizebox{\textwidth}{!}{%
	\centering
        \ra{1.2}
	\begin{tabular}{lccccccccccc}
				\toprule
				\textbf{Model} & \textbf{Education} & \textbf{Food} & \textbf{Games} & \textbf{Health} & \textbf{Lifestyle} & \textbf{Movie} & \textbf{Music} & \textbf{News} & \textbf{Novelty} & \textbf{Smart Home} \\ 
				\hline
	            \ModelName{Joint-SF-IC} & 83.83 & 71.29 & 85.29 & 74.18 & 73.68 & 70.45 & 78.37 & 71.15 & 74.12 & 76.07 \\
				\hdashline
				\ModelName{Parallel[Univ]} & 84.75 & \textbf{76.54} & \textbf{86.43} & \textbf{74.50} & 71.85 & 72.32 & 78.46 & 70.53 & \textbf{75.22} & 76.48 \\
				\ModelName{Parallel[Univ+Task]} & 84.59 & 73.22 & 85.80 & 69.60 & 76.76 & 75.43 & 78.38 & 70.67 & 74.60 & 76.97 \\
				\ModelName{Parallel[Univ+Group+Task]} & 84.41 & 76.43 & 85.68 & 70.81 & \textbf{78.24} & \textbf{76.74} & 78.63 & 72.33 & 74.52 & 77.22 \\
				\hdashline
	            \ModelName{Serial} & 84.74 & 71.79 & 85.42 & 73.02 & 72.17 & 73.56 & 79.30 & 71.90 & 74.37 & 77.56 \\
				\ModelName{Serial+Highway} & \textbf{85.20} & 74.13 & 85.78 & 71.58 & 73.43 & 74.29 & \textbf{80.12} & 71.40 & 74.23 & \textbf{77.75} \\
	            \ModelName{Serial+Highway+Swap} & 84.93 & 74.87 & \textbf{86.35} & 72.38 & 72.02 & 72.09 & 78.86 & \textbf{72.12} & 74.49 & 77.69 \\
	            \bottomrule
	        \end{tabular}
        }
	\caption{Slot F1 on different groups of the Alexa dataset.}
	\label{table:result_alexa_slot_group}
\end{table*}

\subsection{Benchmark data}\label{sec:benchmark-data}
Table \ref{table:result_benchmark_1} shows
the results on ATIS and the original version of the Snips dataset (as shown in
Table \ref{table:benchmark data}). In the first four lines, ATIS and Snips are trained separately. In the last two lines (\ModelNameText{Parallel}), they are treated as two tasks in the MTL setup. There are no task groups in this particular experiment, as each utterance belongs to either ATIS or Snips, and all utterances belong to the task universe.
The \ModelNameText{Joint-SF-IC} architecture with CRF
layer performs better than all the three baseline models in terms of
all evaluation metrics on both datasets, even after removing the
slot-gate \cite{goo2018slot} and attention
\cite{liu2016attention}. Learning universe features across both the
datasets in addition to the task features help ATIS while performance
on Snips degrades. This might be due to the fact that Snips is a multi-domain dataset, which in turn motivates us to split the Snips dataset (as shown in Table \ref{table:snips split}), so that the tasks in each domain (i.e., task group) may share features separately. 

Table \ref{table:result_benchmark_2} shows results on ATIS and our split version of Snips. We now have four tasks: ATIS, Snips-location, Snips-music, and Snips-creative. \ModelNameText{Joint-SF-IC} is our baseline that treats these four tasks independently. All other models process the four tasks together in the MTL setup. For the models introduced in this paper, we define two task groups: ATIS and Snips-location as one group, and Snips-music and Snips-creative as another. Our models, which use these groups, generally outperform the other MTL models (\ModelNameText{Parallel[Univ]} and \ModelNameText{Parallel[Univ+Task]}); especially the serial MTL architectures perform well.

\subsection{Alexa data}
Table \ref{table:result_alexa} shows the results of the single-domain model and the MTL models on the Alexa dataset. The trend is clearly visible in these results compared to the results on the benchmark data. As Alexa data has more domains, there might not be many features that are common across all the domains. Capturing those features that are only common across a group became possible by incorporating task group encoders. \ModelNameText{Serial+Highway+Swap} yields the best mean intent accuracy. \ModelNameText{Parallel+Univ+Group+Task} and \ModelNameText{Serial+Highway} show statistically indistinguishable results. For slot filling, all MTL architectures achieve competitive results on mean Slot F1. 

Overall, on both benchmark data and Alexa data, our architectures with group encoders show better results than others. Specifically, the serial architecture with highway connections achieves the best mean Slot F1 of 94.8 and 97.2 on Snips-music and Snips-location respectively and median Slot F1 of 81.99 on the Alexa dataset. Swapping its feature hierarchy enhances its intent accuracy to 97.5 on ATIS. It also achieves the best/competitive mean and median values on both SF and IC on the Alexa dataset. This supports our argument that when we try to learn common features across all the domains \cite{liu2017adversarial}, we might miss crucial features that are only present across a group. Capturing those task group features boosts the performance of our unified model on SF and IC. In addition, when we attempt to learn three sets of features -- task, task universe, and task group features -- the serial architecture for feature learning helps. Specifically, when we have datasets from many domains, learning task features in the first stage and common features, i.e., task universe and task group features, in the second stage yields the best results. This difference is more clearly visible in the results of the large-scale Alexa data than that of the small-scale benchmark dataset.

\section{Result Analysis}

To further investigate the performance of different architectures, we present the intent accuracy and slot F1 values on different groups of Alexa utterances in Tables~\ref{table:result_alexa_intent_group} and \ref{table:result_alexa_slot_group}. For intent classification, \ModelNameText{Serial+Highway+Swap} achieves the best results on six domains, and \ModelNameText{Parallel[Univ]} achieves the best results on the movie and news domains. Such a finding helps explain the reason why \ModelNameText{Parallel[Univ]} is significantly indistinguishable from \ModelNameText{Serial+Highway+Swap} on the Alexa dataset, which is shown in Table \ref{table:result_alexa}. \ModelNameText{Parallel[Univ]} outperforms MTL with group encoders when there is more information shared across domains. Examples of similar training utterances in different domains are ``go back eight hour'' and ``rewind for eighty five hour'' in a \emph{News} skill; ``to rewind the Netflix'' in a \emph{Smart Home} skill; and ``rewind nine minutes'' in a \emph{Music} skill. The diverse utterance context in different domains could be learned through the universe encoder, which helps to improve the intent accuracy for these skills.

For the slot filling, each domain favors one of the four MTL architectures including \ModelNameText{Parallel[Univ]}, \ModelNameText{Parallel[Univ+Group+Task]}, \ModelNameText{Serial+Highway}, and \ModelNameText{Serial+Highway+Swap}. Such a finding is consistent with the statistically indistinguishable performance between different MTL architectures shown in Table \ref{table:result_alexa}. Tables~\ref{table:incorrect} and \ref{table:correct} show a few utterances from different datasets in the \emph{Smart Home} category that are correctly predicted after learning task group features. General words like \textit{sixty}, \textit{eight}, \textit{alarm} can have different slot types across different datasets. Learning features of the \textit{\mbox{Smart Home}} category helps overcome such conflicts. However, a word in different tasks in the same domain can still have different slot types. For example, the first two utterances in Table~\ref{table:sample-utterances}, which are picked from the \textit{Smart Home} domain, have different slot types \textit{Name} and \textit{Channel} for the word \textit{one}. In such cases, there is no guarantee that learning group features can overcome the conflicts. This might be due to the fact that the groups are predefined and they do not always represent the real task structure. To tackle this issue, learning task structures with features jointly \cite{zhang2017aspect} rather than relying on predefined task groups, would be a future direction. In our experimental settings, all the universe, task, and  task group encoders are instantiated with Bi-LSTM. An interesting area for future experimentation is to streamline the encoders, e.g., adding additional bits to the inputs to the task encoder to indicate the task and group information, which is similar to the idea of using a special token as a representation of the language in a multilingual machine translation system \cite{johnson2017google}.

\begin{table}[!htbp]
\centering
\ra{1.2}
\resizebox{\columnwidth}{!}{%
\centering
\begin{tabular}{l}
				\toprule
\multicolumn{1}{l}{\textbf{Utterance}}                           \\ \hline
				turn\Tag{Other}  to\Tag{Other}  channel\Tag{Other} sixty\TagRed{Name}  eight\TagRed{Name}      \\  
				go\Tag{Other}  to\Tag{Other}  channel\Tag{Other}  sixty\TagRed{VolumeLevel}  seven\TagRed{Name}  \\ 
				turn\Tag{Other}  the\Tag{article}  alarm\TagRed{device}  away\TagRed{device}               \\
				\bottomrule
\end{tabular}
}
\caption{Predictions (incorrect predictions are marked in red) from \textit{Smart Home} domain by the \ModelNameText{Parallel[Univ+Task]} architecture.}
\label{table:incorrect}
\end{table}

\begin{table}[!htbp]
\centering
\ra{1.2}
\resizebox{\columnwidth}{!}{%
\centering
\begin{tabular}{l}
				\toprule
\multicolumn{1}{l}{\textbf{Utterance}}                           \\ \hline
				turn\Tag{Other}  to\Tag{Other}  channel\Tag{Other}  sixty\TagGreen{Channel}  eight\TagGreen{Channel} \\ 
				go\Tag{Other}  to\Tag{Other}  channel\Tag{Other}  sixty\TagGreen{Channel}  seven\TagGreen{Channel}   \\ 
				turn\Tag{Other}  the\Tag{article}  alarm\TagGreen{security\_system} away\TagGreen{type}  \\
				\bottomrule
\end{tabular}
}
\caption{Predictions (correct predictions are marked in green) from \textit{Smart Home} domain by the \ModelNameText{Serial+Highway+Swap} architecture.}
\label{table:correct}
\end{table}

\begin{table}[!htbp]
\centering
\ra{1.2}
\resizebox{\columnwidth}{!}{%
\centering
\begin{tabular}{l}\toprule
\multicolumn{1}{l}{\textbf{Utterance}}                           \\ \hline
tune\Tag{Other}  to\Tag{Other}  the\Tag{Other}  bbc\Tag{Name}  one\TagBlue{Name}  station\Tag{Other} \\ 
change\Tag{Other}  to\Tag{Other}  channel\Tag{Other}  one\TagBlue{Channel}                 \\ 
score\Tag{Other}  sixty\Tag{Number}  one\TagBlue{Number}                              \\ 
four\Tag{Answer}  one\TagBlue{Answer}                                            \\ \bottomrule 
\end{tabular}
}
\caption{Training samples from different domains with different slot types for the word \textit{one} (highlighted in blue).}
\label{table:sample-utterances}
\end{table}

\section{Related Work}

Multi-task learning (MTL) aims to learn multiple related tasks from data simultaneously to improve the predictive performance compared with learning independent models. Various MTL models have been developed based on the assumption that all tasks are related \cite{argyriou2007multi,negahban2008joint,jalali2010dirty}. To tackle the problem that task structure is usually unclear, \newcite{evgeniou2004regularized} extended support vector machines for single-task learning in a multi-task scenario by penalizing models if they are too far from a mean model. \newcite{xue2007multi} introduced a Dirichlet process prior to automatically identify subgroups of related tasks. \newcite{passos2012flexible} developed a nonparametric Bayesian model to learn task subspaces and features jointly.

On the other hand, with the advent of deep learning, MTL with deep neural networks has been successfully applied to different applications \cite{zhang2018multi,masumura2018adversarial,fares2018transfer,guo2018soft}. 
Recent work on multi-task learning considers different sharing structures, e.g., only sharing at lower layers \cite{sogaard2016deep} and introduces private and shared subspaces \cite{liu2016recurrent,liu2017adversarial}. \newcite{liu2017adversarial} incorporated adversarial loss and orthogonality constraints into the overall training object, which helps in learning task-specific and task-invariant features in a non-redundant way. However, they do not explore task structures, which can contain crucial features only present within groups of tasks. Our work encodes task structure information in deep neural architectures. 

\section{Conclusions}
We proposed a series of end-to-end multi-task learning architectures, in which task, task group and task universe features are learned non-redundantly. We further explored learning these features in parallel and serial MTL architectures. Our MTL models obtain state-of-the-art performance on the ATIS and Snips datasets for intent classification and slot filling. Experimental results on a large-scale Alexa dataset show the effectiveness of adding task group encoders into both parallel and serial MTL networks.

\section*{Acknowledgments}

We thank Lambert Mathias for providing insightful feedback and Sandesh Swamy for preparing the Alexa test dataset. We also thank our team members as well as the anonymous reviewers for their valuable comments.

\bibliography{acl2018_short}
\bibliographystyle{acl_natbib}

\end{document}